\documentclass[conference]{IEEEtran}

\usepackage{changes}
\usepackage{amsmath}
\definechangesauthor[color=cyan]{JJ}
\definechangesauthor[color=blue]{CG}
\usepackage{multirow}
\usepackage{graphicx}
\usepackage{hyperref}
\usepackage[11pt]{moresize}
\usepackage{todonotes}

\usepackage{subcaption}

\begin{document}
\title{DELAUNAY: a dataset of abstract art for psychophysical and machine learning research}

\author{\IEEEauthorblockN{ \\Camille Gontier}
\IEEEauthorblockA{Department of Physiology\\University of Bern, Switzerland\\\href{mailto:camille.gontier@unibe.ch}{camille.gontier@unibe.ch}}
\and
\IEEEauthorblockN{ \\Jakob Jordan}
\IEEEauthorblockA{Department of Physiology\\University of Bern, Switzerland\\\href{mailto:jakob.jordan@unibe.ch}{jakob.jordan@unibe.ch}}
 
\and
\IEEEauthorblockN{Mihai A. Petrovici}
\IEEEauthorblockA{Department of Physiology\\University of Bern, Switzerland;\\Kirchhoff-Institute for Physics \\Heidelberg University, Germany \\ \href{mailto:mihai.petrovici@unibe.ch}{mihai.petrovici@unibe.ch}}
} 

\maketitle

\begin{abstract}
	Image datasets are commonly used in psychophysical experiments and in machine learning research.
	Most publicly available datasets are comprised of images of realistic and natural objects.
	However, while typical machine learning models lack any domain specific knowledge about natural objects, humans can leverage prior experience for such data, making comparisons between artificial and natural learning challenging.
	Here, we introduce DELAUNAY,
	a dataset of abstract paintings and non-figurative art objects labelled by the artists' names.
	This dataset provides a middle ground between natural images and artificial patterns and can thus be used in a variety of contexts, for example to investigate the sample efficiency of humans and artificial neural networks.
	Finally, we train an off-the-shelf convolutional neural network on DELAUNAY, highlighting several of its intriguing features.
\end{abstract}

\IEEEpeerreviewmaketitle

\section{Introduction}

Deep Neural Networks (DNNs) have for many years demonstrated human and even super-human performance in many different tasks (see \cite{lecun2015deep} for a review).
One of their most famous achievements is super-human natural image classification: during the 2017 edition of the ImageNet Large Scale Visual Recognition Challenge (ILSVRC), the winning team achieved a classification error of $2.251\%$, using a network trained on a subset of ImageNet containing 1000 categories and 1.2 million images (\cite{ILSVRC15,Hu_2018_CVPR}).
In comparison, the estimated human classification error on the same task is $5.1\%$ (\cite{ILSVRC,russakovsky2015imagenet}).
They also perform extremely well in other vision-related tasks, such as boundary detection, semantic segmentation, semantic boundaries, surface normals, saliency, human anatomy, and object detection (\cite{yuille2018deep}).
DNNs and the human visual system share architectural similarities, but their link may be even deeper than merely structural, as the activity of different artificial cells in a DNNs can be mapped to, and subsequently used to predict, the recorded activity of cells in a living subject (\cite{yamins2014performance,bashivan2019neural}).
Furthermore, recent years have witnessed the emergence of a multitude of models linking learning in DNNs to synaptic plasticity in cortex (\cite{guerguiev2017towards,whittington2017approximation,sacramento2018dendritic,whittington2019theories,lillicrap2020backpropagation,sun2021bursting,haider2021latent}).

\begin{figure*}[t]
    \centering
	\includegraphics[width=0.99\textwidth]{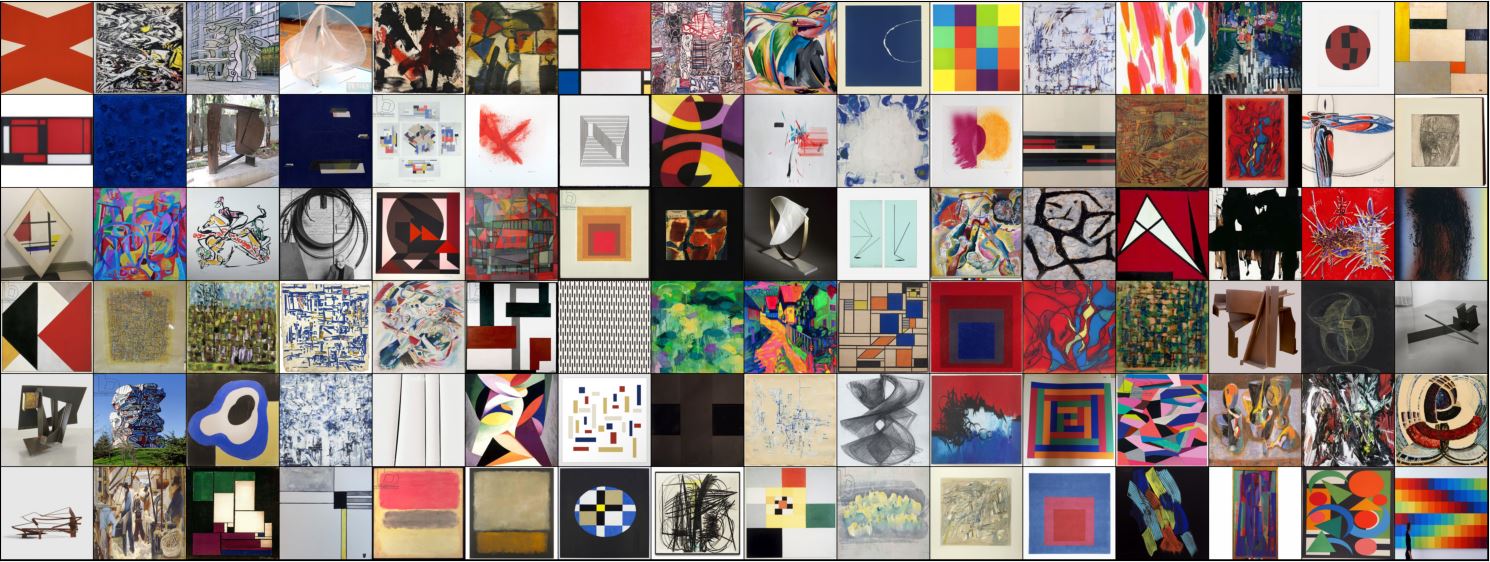}
	\caption{{\bf Samples from DELAUNAY.}
	DELAUNAY consists of images of abstract artwork from a variety of different artists.
	Here, 96 random samples across all artists are shown.
	Note the diverse non-figurative properties of images.}
	\label{fig:examples}
\end{figure*}

Despite these successes, a number of critical shortcomings of these systems have been observed over the recent years.
For example, DNNs are notorious for the large amount of labeled training data required (\cite{huttunen2016car}).
Humans, in contrast, are extremely efficient learners, able to learn new categories from very few samples.
Furthermore, human perception is much more robust to rotations, occlusions or even abstraction of image content.
These observations suggest that DNNs and humans leverage fundamentally different strategies for learning.

The learning performance of a system is intimately linked to its prior knowledge of the task domain and inductive biases (\cite{sinz2019engineering}).
Thus, one may hypothesize that the difference in sample efficiency between neural networks and human subjects may significantly depends on the considered task.
For example, classification tasks involving images with natural structures may be learnt faster by humans, by exploiting biases and priors which were developed over evolutionary timescales and acquired over the lifetime of an organism.
However, classifying artificial inputs, such as pseudo-random patterns or QR-codes may take humans longer to learn, effectively resulting in memorization, leading to lower sample efficiency and poor extrapolation.

A number of researchers have started to compare natural and artificial learning leveraging psychophsyics and machine learning with the twin goals of furthering our understanding of brain function and improving artificial intelligence (\cite{broker2021unsupervised,geirhos2019generalisation,lake2017building,tenenbaum2011grow}).
In a similar spirit, to measure how much prior knowledge and inductive biases contribute to the learning speed of humans, one may design psychophysical experiments comparing the sample efficiency of DNNs and human subjects on visual classification tasks in which the statistical similarity of the datasets to natural images is controlled:
\begin{enumerate}
    \item A task with realistically structured objects, for example ImageNet (\cite{russakovsky2015imagenet}). 
    \item A task with images of unusual structures, for example abstract art.
    \item A task with completely unstructured images, for example QR codes, or shuffled MNIST (\cite{goodfellow2013empirical,srivastava2013compete}). 
\end{enumerate}

In addition, the second task, i.e., the classification of abstract art, could consider two different subject groups, consisting of naive viewers and art experts, respectively.
Both the naive group in task 2 and all subjects in task 3 can be considered lacking prior task-related knowledge which should lead to a significant drop in sample efficiency.

If these expectations turn out to be correct, i.e., if we observe a significant difference in the relative sample efficiency of humans compared to DNNs depending on the statistical similarity of the dataset to natural images, this would provide support for the hypothesis that fast natural and artificial learning heavily relies on prior domain knowledge, as speculated for example by \cite{zador2019critique}.

Our main contribution here is to provide a new dataset consisting of images of abstract art (as opposed to other databases of paintings: \cite{lecoutre2017recognizing,khan2014painting}), suitable for psychophysical experiments and machine learning research.


\section{Dataset}	

DELAUNAY (Dataset for Experiments on Learning with Abstract and non-figurative art for Neural networks and Artificial intelligence) is named after artists Sonia \cite{enwiki:1040963854} and Robert Delaunay \cite{enwiki:1052806387}.

Several museums and other institutions worldwide offer remote access to large databases, such as the Solomon R. Guggenheim Museum in New-York \cite{Guggenheim}, the MET \cite{MET}, the Bibliothèque Nationale de France (BNF) through its open-access online tool Gallica \cite{wiki:gallica}, the digital collections of the Library of Congress \cite{libraryofcongress}, the French Réunion des Musées Nationaux - Grand Palais (Rmn-GP) \cite{rmn}, the online library of the Institut National d'Histoire de l'Art (INHA) \cite{inha}, the Bridgeman Art Library \cite{bridgeman}, the National Portrait Gallery \cite{npg}, as well as the Alamy and Getty Images photo libraries \cite{wiki:alamy}.

We leveraged these online databases to construct DELAUNAY.
First, we selected 53 artists well known for their abstract art (see the full list in Annex). Second, we scraped the aforementioned databases for images of their artworks.
Finally, we removed false positives (e.g., photographs of the artists), duplicates, and cropped images containing texts and/or other artifacts.

The final dataset comprises $11,503$ samples across $53$ classes, i.e., artists (mean number of samples per artist: $217.04$; standard deviation: $58.55$), along with their source URLs.
These samples are split between a training set of 9202 images and a test set of 2301 images.
Due to the heterogeneous nature of sources, images vary significantly in their resolution, from $80\text{px} \times 80\text{px}$ for the smallest sample to $3365\text{px} \times 4299\text{px}$ for the largest.
A random subset of samples illustrating their non-figurative (not representing a natural object) nature and high diversity is shown in Figure \ref{fig:examples}.

\section{Training results}\label{sec:training}

To illustrate some of the unique features of this new dataset, we train a standard convolutional neural network (CNN) to classify samples according to their authoring artist.
We used a similar approach as in \cite{geirhos2019generalisation}: images were resized to $256 \times 256$ pixels (not preserving their aspect ratio) and classified using the ResNet152 architecture (\cite{he2016deep}) distributed as part of PyTorch (\cite{NEURIPS2019_9015}).
Parameters were randomly initialized according to PyTorch's defaults.
We trained the network using a cross entropy loss with the ADAM optimizer (\cite{kingma2014adam}) with a learning rate $\gamma = 0.003$, weight decay parameter $\lambda = 0.003$, and batch size 20.

\begin{figure}[tbp]
    \centering
	\includegraphics[width=0.49\textwidth]{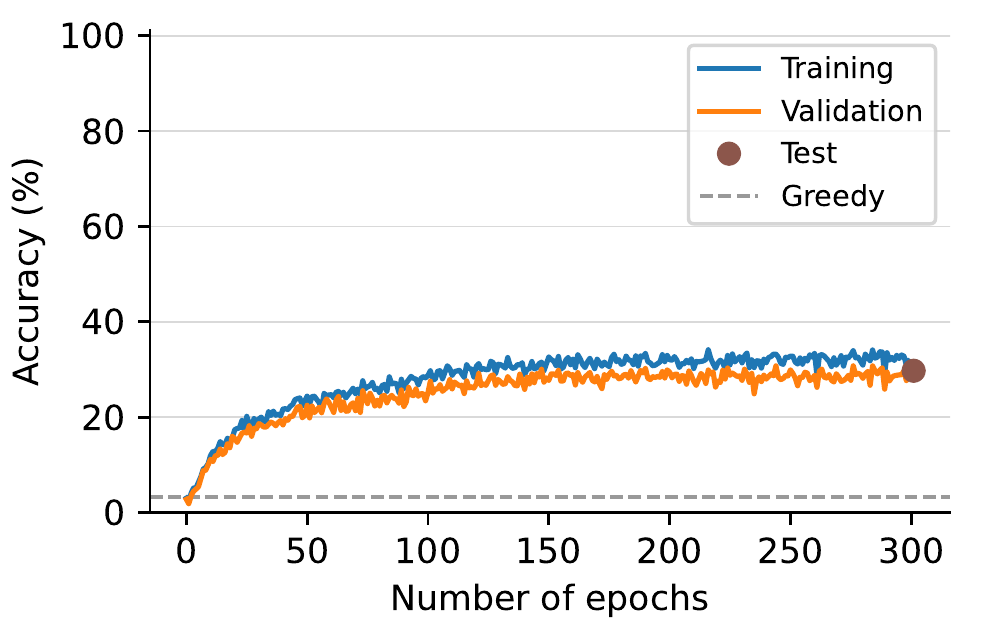}
	\caption{{\bf Training results of a ResNet152 on DELAUNAY.}
	Curves represent training and validation accuracy over epochs.
	Marker indicates test accuracy.
	Dashed line represent performance of a greedy classifier with output fixed to the most common class.
	For training details see Section \ref{sec:training}.}\label{fig:results}
\end{figure}

Both validation and test accuracies are significantly higher (respectively $29.46 \%$ and $29.73 \%$) than the accuracy expected from a greedy classifier associating all images to the class having the highest sample count ($3.28 \%$) (Figure \ref{fig:results}).

%
%

\begin{figure*}[tbp]
    \centering
    \includegraphics[width=1.0\textwidth]{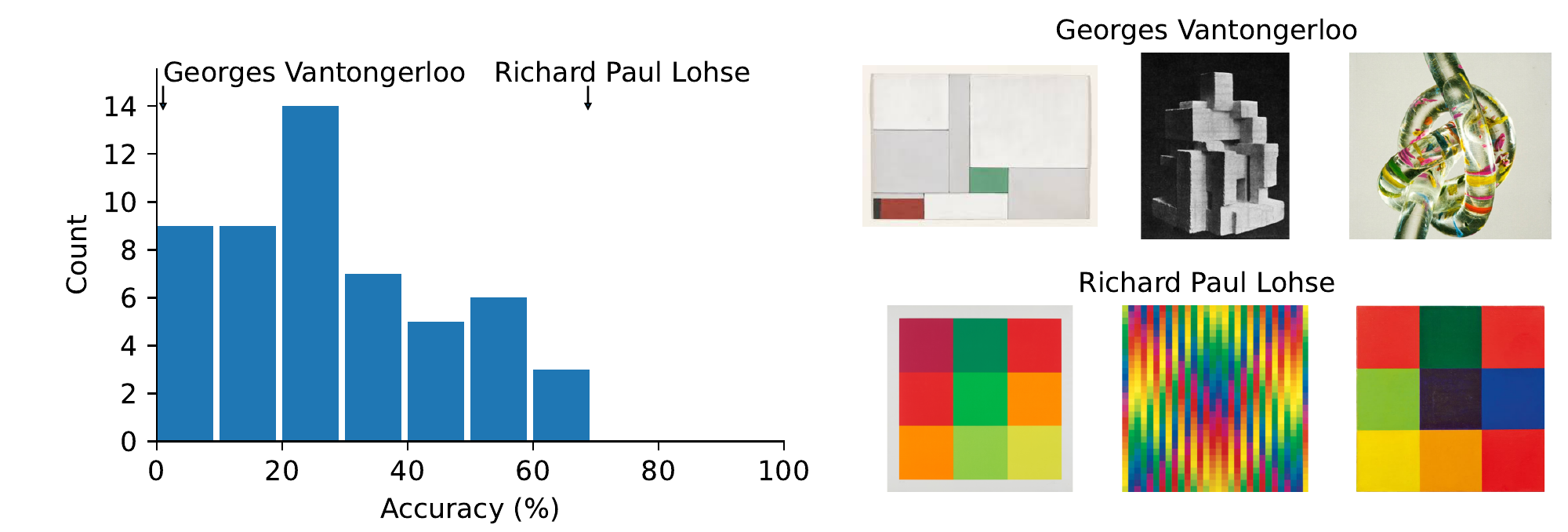}

    \caption{{\bf High intra-class variability in DELAUNAY leads to low test accuracies for some classes.}
    Left: Histogram over accuracies for all 53 classes in the test set.
    Highlighted are two artists with low and high test accuracies, respectively.
    Right: Example of works from Georges Vantongerloo (top) and Richard Paul Lohse (bottom).
    Note the high and low variability of samples within these classes, respectively.}\label{fig:comparison_artists}
\end{figure*}

\begin{figure*}[tbp]
    \centering
	\includegraphics[width=1.\textwidth]{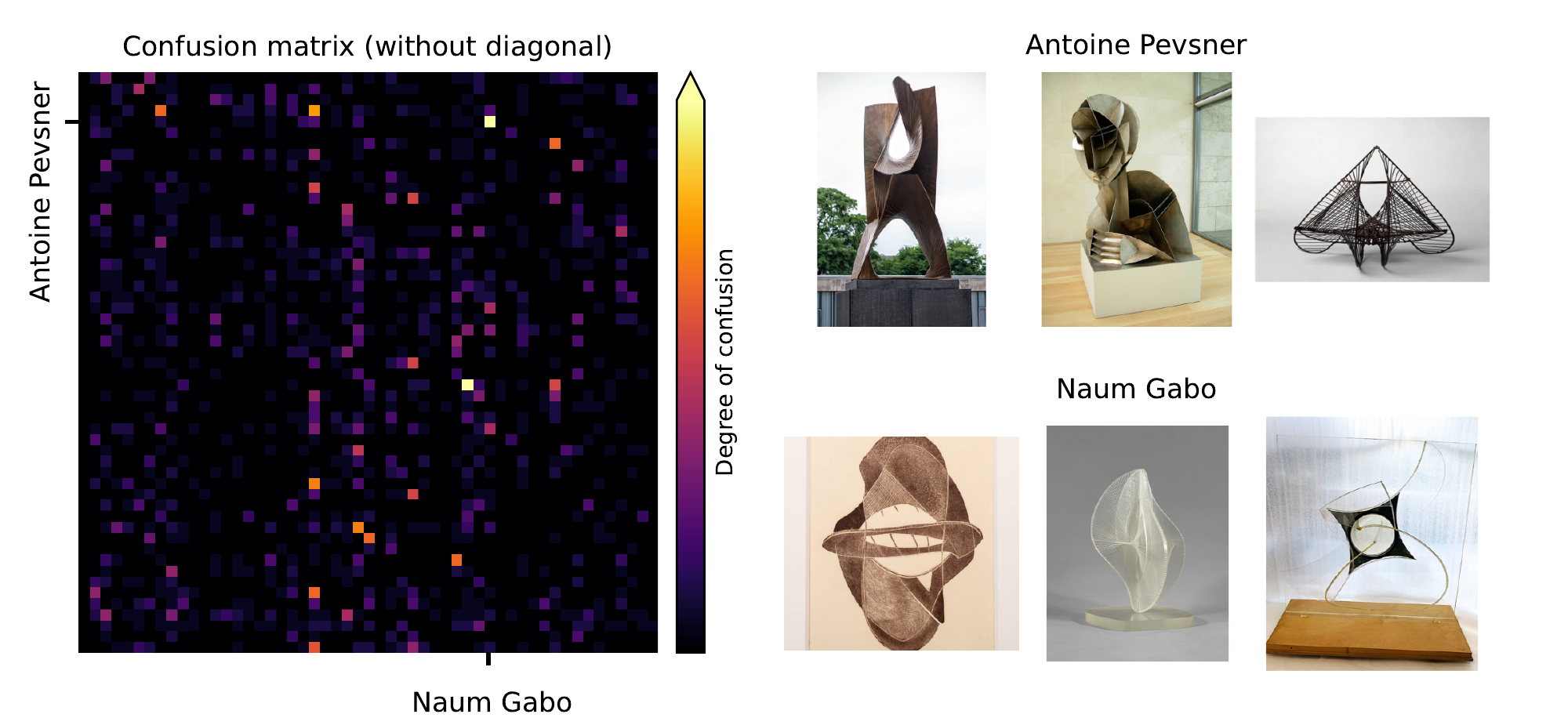}
	\caption{{\bf Low inter-class variability leads to high degree of confusion for some class pairs.}
	Left: Confusion matrix for all 53 classes in the test set.
	Diagonal entries were removed.
	Right: Example of works from Naum Gabo (top) and Antoine Pevsner (bottom).
	Note the high similarity of samples from these different classes.}\label{fig:confusion}
\end{figure*}

To investigate these results further, we consider class-level error metrics.
Percentages of correct predictions for all 53 classes in the test set range from $0.0\%$ to $68.75\%$ (Figure \ref{fig:comparison_artists}).
This wide distribution suggests that while some artists can be easily recognized by the trained CNN, some seem to withstand meaningful extrapolation from the training to the test set.
Indeed, artists with low accuracies appear to have a multimodal, eclectic style in contrast to other artists who maintain a more systematic style throughout their works (compare for example Georges Vantongerloo to Richard Paul Lohse, Figure \ref{fig:comparison_artists}).
Furthermore, the confusion matrix of the test dataset contains several significant off-diagonal entries (Figure \ref{fig:confusion}), indicating that some artists are particularly often confused by the trained CNN, likely due to a high similarity of their works.
From an historical perspective, such a high degree of similarity is not surprising: rather than working in isolation, artists are often significantly influenced by specific artistic movements and their personal environment such as close friends and family.
For example, we observe a high degree of confusion between the works of Naum Gabo and his brother Antoine Pevsner, which indeed can be traced back to a high degree of similarity samples from these classes (Figure \ref{fig:confusion}).

\section{Discussion/Conclusion}

We have introduced DELAUNAY, a dataset of images of abstract and non-figurative artworks from 53 different artists.
It provides a middle ground between natural images typically used in machine learning research and unnatural, structureless patterns at the opposite side of the spectrum.
We believe the unique properties of this dataset make it useful for both machine learning as well as psychophsyics research, for example to investigate the hypothesis that sample efficiency scales inversely with the statistical similarity of samples to natural images for humans but not for DNNs.

Here we illustrated two intriguing properties of the dataset which make it challenging for classical deep-learning approaches: first, the intra-class variance for some classes is large, and second the inter-class variance for some classes is (relatively) small.
We believe that addressing these challenges, supported by insights about human strategies obtained from psychophysical experiments, is a fruitful direction to both further understanding brain function as well as developing new machine learning methods.
We are excited about seeing the community put this dataset to creative use.

\section*{Code and data availability}

Link to the dataset (including the original URLs for all samples), as well as scripts used for the creation of the dataset, for training the CNN, and analysis of the results are available from \href{https://github.com/camillegontier/DELAUNAY_dataset}{https://github.com/camillegontier/DELAUNAY\_dataset}.

\section*{Acknowledgments}

Calculations were performed on UBELIX (\url{http://www.id.unibe.ch/hpc}), the HPC cluster at the University of Bern.
We thank Jean-Pascal Pfister, Walter Senn, and Manon Scoubeau for fruitful discussions.


\section*{Annex}

{\ssmall Artists included in the dataset:
Josef Albers,
Jean Arp,
Olle Bærtling,
Jean Bazaine,
Étienne Béothy,
Roger Bissière,
Anthony Caro,
Jean Degottex,
Sonia and Robert Delaunay,
César Domela,
Jean Dubuffet,
Jean Fautrier,
Lucio Fontana,
Sam Francis,
Otto Freundlich,
Naum Gabo,
Léon Gischia,
Jean Gorin,
Hans Hartung,
Auguste Herbin,
Vassily Kandinsky,
Ellsworth Kelly,
Yves Klein,
Franz Kline,
František Kupka,
Charles Lapicque,
Berto Lardera,
Fernand Léger,
Richard Paul Lohse,
Morris Louis,
Alberto Magnelli, 
Alfred Manessier,
Georges Mathieu,
Joan Mitchell,
László Moholy-Nagy,
Piet Mondrian, 
François Morellet,
Aurélie Nemours,
Kenneth Noland,
Antoine Pevsner,
Leon Polk Smith,
Ad Reinhardt,
Mark Rothko,
Gustave Singier,
Pierre Soulages, 
Sophie Taeuber-Arp,
Pierre Tal Coat,
Theo van Doesburg,
Georges Vantongerloo,
Victor Vasarely,
Emilio Vedova,
Maria Helena Vieira da Silva,
Charmion Von Wiegand.}
    

\bibliography{covert}{}

\begin{thebibliography}{10}

\bibitem{lecun2015deep}
Y.~LeCun, Y.~Bengio, and G.~Hinton, ``Deep learning,'' {\em Nature}, vol.~521,
  no.~7553, pp.~436--444, 2015.

\bibitem{ILSVRC15}
O.~Russakovsky, J.~Deng, H.~Su, J.~Krause, S.~Satheesh, S.~Ma, Z.~Huang,
  A.~Karpathy, A.~Khosla, M.~Bernstein, A.~C. Berg, and L.~Fei-Fei, ``{ImageNet
  Large Scale Visual Recognition Challenge},'' {\em International Journal of
  Computer Vision (IJCV)}, vol.~115, no.~3, pp.~211--252, 2015.

\bibitem{Hu_2018_CVPR}
J.~Hu, L.~Shen, and G.~Sun, ``Squeeze-and-excitation networks,'' in {\em The
  IEEE Conference on Computer Vision and Pattern Recognition (CVPR)}, June
  2018.

\bibitem{ILSVRC}
``{What I learned from competing against a ConvNet on ImageNet}.''
  \url{https://karpathy.github.io/2014/09/02/what-i-learned-from-competing-against-a-convnet-on-imagenet/}.
\newblock [Online; accessed 02-April-2021].

\bibitem{russakovsky2015imagenet}
O.~Russakovsky, J.~Deng, H.~Su, J.~Krause, S.~Satheesh, S.~Ma, Z.~Huang,
  A.~Karpathy, A.~Khosla, M.~Bernstein, {\em et~al.}, ``Imagenet large scale
  visual recognition challenge,'' {\em International Journal of Computer
  Vision}, vol.~115, no.~3, pp.~211--252, 2015.

\bibitem{yuille2018deep}
A.~L. Yuille and C.~Liu, ``Deep nets: What have they ever done for vision?,''
  {\em arXiv preprint arXiv:1805.04025}, 2018.

\bibitem{yamins2014performance}
D.~L. Yamins, H.~Hong, C.~F. Cadieu, E.~A. Solomon, D.~Seibert, and J.~J.
  DiCarlo, ``Performance-optimized hierarchical models predict neural responses
  in higher visual cortex,'' {\em Proceedings of the National Academy of
  Sciences}, vol.~111, no.~23, pp.~8619--8624, 2014.

\bibitem{bashivan2019neural}
P.~Bashivan, K.~Kar, and J.~J. DiCarlo, ``Neural population control via deep
  image synthesis,'' {\em Science}, vol.~364, no.~6439, p.~eaav9436, 2019.

\bibitem{guerguiev2017towards}
J.~Guerguiev, T.~P. Lillicrap, and B.~A. Richards, ``Towards deep learning with
  segregated dendrites,'' {\em eLife}, vol.~6, p.~e22901, 2017.

\bibitem{whittington2017approximation}
J.~C. Whittington and R.~Bogacz, ``An approximation of the error
  backpropagation algorithm in a predictive coding network with local hebbian
  synaptic plasticity,'' {\em Neural Computation}, vol.~29, no.~5,
  pp.~1229--1262, 2017.

\bibitem{sacramento2018dendritic}
J.~Sacramento, R.~Ponte~Costa, Y.~Bengio, and W.~Senn, ``Dendritic cortical
  microcircuits approximate the backpropagation algorithm,'' {\em Advances in
  Neural Information Processing Systems}, vol.~31, pp.~8721--8732, 2018.

\bibitem{whittington2019theories}
J.~C. Whittington and R.~Bogacz, ``Theories of error back-propagation in the
  brain,'' {\em Trends in Cognitive Sciences}, vol.~23, no.~3, pp.~235--250,
  2019.

\bibitem{lillicrap2020backpropagation}
T.~P. Lillicrap, A.~Santoro, L.~Marris, C.~J. Akerman, and G.~Hinton,
  ``Backpropagation and the brain,'' {\em Nature Reviews Neuroscience},
  vol.~21, no.~6, pp.~335--346, 2020.

\bibitem{sun2021bursting}
W.~Sun, X.~Zhao, and N.~Spruston, ``Bursting potentiates the neuro--ai
  connection,'' {\em Nature Neuroscience}, pp.~1--3, 2021.

\bibitem{haider2021latent}
P.~Haider, B.~Ellenberger, L.~Kriener, J.~Jordan, W.~Senn, and M.~Petrovici,
  ``Latent equilibrium: Arbitrarily fast computation with arbitrarily slow
  neurons,'' {\em Advances in Neural Information Processing Systems}, vol.~34,
  2021.

\bibitem{huttunen2016car}
H.~Huttunen, F.~S. Yancheshmeh, and K.~Chen, ``Car type recognition with deep
  neural networks,'' in {\em 2016 IEEE Intelligent Vehicles Symposium (IV)},
  pp.~1115--1120, IEEE, 2016.

\bibitem{sinz2019engineering}
F.~H. Sinz, X.~Pitkow, J.~Reimer, M.~Bethge, and A.~S. Tolias, ``Engineering a
  less artificial intelligence,'' {\em Neuron}, vol.~103, no.~6, pp.~967--979,
  2019.

\bibitem{broker2021unsupervised}
F.~Br{\"o}ker, B.~C. Love, and P.~Dayan, ``When unsupervised training benefits
  category learning,'' 2021.

\bibitem{geirhos2019generalisation}
R.~Geirhos, C.~Medina~Temme, J.~Rauber, H.~Sch{\"u}tt, M.~Bethge, and
  F.~Wichmann, ``Generalisation in humans and deep neural networks,'' in {\em
  Thirty-second Annual Conference on Neural Information Processing Systems 2018
  (NeurIPS 2018)}, pp.~7549--7561, Curran, 2019.

\bibitem{lake2017building}
B.~M. Lake, T.~D. Ullman, J.~B. Tenenbaum, and S.~J. Gershman, ``Building
  machines that learn and think like people,'' {\em Behavioral and Brain
  Sciences}, vol.~40, 2017.

\bibitem{tenenbaum2011grow}
J.~B. Tenenbaum, C.~Kemp, T.~L. Griffiths, and N.~D. Goodman, ``How to grow a
  mind: Statistics, structure, and abstraction,'' {\em science}, vol.~331,
  no.~6022, pp.~1279--1285, 2011.

\bibitem{goodfellow2013empirical}
I.~J. Goodfellow, M.~Mirza, D.~Xiao, A.~Courville, and Y.~Bengio, ``An
  empirical investigation of catastrophic forgetting in gradient-based neural
  networks,'' {\em arXiv preprint arXiv:1312.6211}, 2013.

\bibitem{srivastava2013compete}
R.~K. Srivastava, J.~Masci, S.~Kazerounian, F.~J. Gomez, and J.~Schmidhuber,
  ``Compete to compute.,'' in {\em NIPS}, pp.~2310--2318, Citeseer, 2013.

\bibitem{zador2019critique}
A.~M. Zador, ``A critique of pure learning and what artificial neural networks
  can learn from animal brains,'' {\em Nature Communications}, vol.~10, no.~1,
  pp.~1--7, 2019.

\bibitem{lecoutre2017recognizing}
A.~Lecoutre, B.~Negrevergne, and F.~Yger, ``Recognizing art style automatically
  in painting with deep learning,'' in {\em Asian Conference on Machine
  Learning}, pp.~327--342, PMLR, 2017.

\bibitem{khan2014painting}
F.~S. Khan, S.~Beigpour, J.~Van~de Weijer, and M.~Felsberg, ``Painting-91: a
  large scale database for computational painting categorization,'' {\em
  Machine Vision and Applications}, vol.~25, no.~6, pp.~1385--1397, 2014.

\bibitem{enwiki:1040963854}
{Wikipedia contributors}, ``{Sonia Delaunay} --- {Wikipedia}{,} the free
  encyclopedia,'' 2021.
\newblock [Online; accessed 15-November-2021].

\bibitem{enwiki:1052806387}
{Wikipedia contributors}, ``{Robert Delaunay} --- {Wikipedia}{,} the free
  encyclopedia,'' 2021.
\newblock [Online; accessed 15-November-2021].

\bibitem{Guggenheim}
``{The Solomon R. Guggenheim Museum}.''
  \url{https://archive.org/details/guggenheimmuseum}.
\newblock [Online; accessed 22-March-2020].

\bibitem{MET}
``{The Met Collection}.''
  \url{https://www.metmuseum.org/art/collection/search#!?searchField=All&showOnly=openAccess&sortBy=Relevance&offset=0&pageSize=0}.
\newblock [Online; accessed 22-March-2020].

\bibitem{wiki:gallica}
Wikipédia, ``Gallica --- wikipédia{,} l'encyclopédie libre,'' 2020.
\newblock [En ligne; Page disponible le 27-janvier-2020].

\bibitem{libraryofcongress}
``{Digital collections, Library of Congress}.''
  \url{https://www.loc.gov/collections/}.
\newblock [Online; accessed 22-March-2020].

\bibitem{rmn}
``{Réunion des Musées Nationaux - Grand Palais (Rmn-GP)}.''
  \url{https://art.rmngp.fr/fr}.
\newblock [Online; accessed 22-March-2020].

\bibitem{inha}
``{Institut National d'Histoire de l'Art}.''
  \url{https://www.inha.fr/fr/bibliotheque/bibliotheque-numerique.html}.
\newblock [Online; accessed 22-March-2020].

\bibitem{bridgeman}
``{Bridgeman Art Library}.'' \url{https://www.bridgemanimages.us/en-US/}.
\newblock [Online; accessed 22-March-2020].

\bibitem{npg}
``{National Portrait Gallery}.'' \url{https://www.npg.org.uk/}.
\newblock [Online; accessed 22-March-2020].

\bibitem{wiki:alamy}
{Wikipedia contributors}, ``Alamy --- {Wikipedia}{,} the free encyclopedia.''
  \url{https://en.wikipedia.org/w/index.php?title=Alamy&oldid=942994711}, 2020.
\newblock [Online; accessed 23-March-2020].

\bibitem{he2016deep}
K.~He, X.~Zhang, S.~Ren, and J.~Sun, ``Deep residual learning for image
  recognition,'' in {\em Proceedings of the IEEE conference on computer vision
  and pattern recognition}, pp.~770--778, 2016.

\bibitem{NEURIPS2019_9015}
A.~Paszke, S.~Gross, F.~Massa, A.~Lerer, J.~Bradbury, G.~Chanan, T.~Killeen,
  Z.~Lin, N.~Gimelshein, L.~Antiga, A.~Desmaison, A.~Kopf, E.~Yang, Z.~DeVito,
  M.~Raison, A.~Tejani, S.~Chilamkurthy, B.~Steiner, L.~Fang, J.~Bai, and
  S.~Chintala, ``Pytorch: An imperative style, high-performance deep learning
  library,'' in {\em Advances in Neural Information Processing Systems 32},
  pp.~8024--8035, Curran Associates, Inc., 2019.

\bibitem{kingma2014adam}
D.~P. Kingma and J.~Ba, ``Adam: A method for stochastic optimization,'' {\em
  arXiv preprint arXiv:1412.6980}, 2014.

\end{thebibliography}
\bibliographystyle{ieeetr}
\vfill

{\ssmall
\begin{center}
\begin{tabular}{ |c|c|c| } 
\hline
Artist & Sample count \\
\hline
Josef Albers & 285 \\
Jean Arp & 337 \\
Olle Bærtling & 151 \\
Jean Bazaine & 207 \\
Étienne Béothy & 207 \\
Roger Bissière & 198 \\
Anthony Caro & 261 \\
Jean Degottex & 212 \\
Sonia and Robert Delaunay & 176 \\
César Domela & 171 \\
Jean Dubuffet & 364 \\
Jean Fautrier & 269 \\
Lucio Fontana & 159 \\
Sam Francis & 280 \\
Otto Freundlich & 181 \\
Naum Gabo & 261 \\
Léon Gischia & 79 \\
Jean Gorin & 208 \\
Hans Hartung & 251 \\
Auguste Herbin & 222 \\
Vassily Kandinsky & 377 \\
Ellsworth Kelly & 216 \\
Yves Klein & 123 \\
Franz Kline & 180 \\
František Kupka & 259 \\
Charles Lapicque & 250 \\
Berto Lardera & 203 \\
Fernand Léger & 169 \\
Richard Paul Lohse & 194 \\
Morris Louis & 225 \\
Alberto Magnelli & 265 \\ 
Alfred Manessier & 283 \\
Georges Mathieu & 199 \\
Joan Mitchell & 91 \\
László Moholy-Nagy & 232 \\
Piet Mondrian & 176 \\
François Morellet & 201 \\
Aurélie Nemours & 190 \\
Kenneth Noland & 229 \\
Antoine Pevsner & 218 \\
Leon Polk Smith & 196 \\
Ad Reinhardt & 167 \\
Mark Rothko & 300 \\
Gustave Singier & 256 \\
Pierre Soulages & 176 \\ 
Sophie Taeuber-Arp & 186 \\
Pierre Tal Coat & 195 \\
Theo van Doesburg & 197 \\
Georges Vantongerloo & 170 \\
Victor Vasarely & 312 \\
Emilio Vedova & 202 \\
Maria Helena Vieira da Silva & 223 \\
Charmion Von Wiegand & 164 \\
\hline
\end{tabular}
\end{center}
}

\end{document}